# Understanding and Improving Deep Neural Network for Activity Recognition

Li Xue, Si Xiandong[†], Nie Lanshun, Li Jiazhen, Ding Renjie, Zhan Dechen, Chu Dianhui
{lixuecs@hit.edu.cn, 15776633420@163.com, rhine.nie@gmail.com }

School of Computer Science and Technology, Harbin Institute of Technology

**Abstract.** Activity recognition has become a popular research branch in the field of pervasive computing in recent years. A large number of experiments can be obtained that activity sensor-based data's characteristic in activity recognition is variety, volume, and velocity. Deep learning technology, together with its various models, is one of the most effective ways of working on activity data. Nevertheless, there is no clear understanding of why it performs so well or how to make it more effective. In order to solve this problem, first, we applied convolution neural network on Human Activity Recognition Using Smart phones Data Set. Second, we realized the visualization of the sensor-based activity's data features extracted from the neural network. Then we had in-depth analysis of the visualization of features, explored the relationship between activity and features, and analyzed how Neural Networks identify activity based on these features. After that, we extracted the significant features related to the activities and sent the features to the DNN-based fusion model, which improved the classification rate to 96.1%. This is the first work to our knowledge that visualizes abstract sensor-based activity data features. Based on the results, the method proposed in the paper promises to realize the accurate classification of sensor-based activity recognition.

## 1 Introduction

With the development of the Internet of Things and big data, human activity recognition (HAR) has attracted much attention as a branch of pervasive computing. Recently, many academic researchers have focused on work that proposes various kinds of new methods to improve recognition performance [1][2][3][4]. HAR can also be seen as a branch of pattern recognition. Moreover, industry professionals have put significant effort into developing friendly sensors and equipment for human use. HAR is studied to solve practical problems, improve the safety and comfort of people's lives in their homes and in their daily common activities, and when combined with many professional fields such as medicine [5][6], health [7][8], and sports [9][10] is an inevitable trend. Of course, these achievements get approval.

In the field of activity recognition, convolution neural network (CNN) as the most basic model of deep learning technology shows excellent performance. In [11], they propose an approach to automatically extract discriminative features for activity recognition. Specifically, CNN-based methods can capture local dependence and a signal's scale invariance. Chen, Y., & Xue, Y constructed a CNN model and modified the convolution kernel to adapt the characteristics of tri-axial acceleration signals [12]. Nine types of sensor information are used to synthesize a signal image for 2D image recognition in [13]. The work in [14] and [15] combines sensor data and time information into a two-dimensional matrix form, similar to single-axis picture information, which preserves sensor correlations and time series information. Human activity data based on sensors has high dimension characteristics and data features are difficult to extract using traditional statistical methods. CNN is suitable for extracting high dimensional data, and the automatic extraction of data features is more suitable for the situation in which the data features of a sensor are not obvious in the activity recognition. Another reason why CNN can extract complex activity recognition data features mainly depends on three important features: local connection, weight sharing, and translation invariance [16]. Therefore, this paper uses CNN to extract high-level data features.

Although CNN performs well, it has some problems: 1) It is hard to understand the internal mechanism of neural network. How neural networks recognize activity based on sensor data? What is the relationship between sensor data and activities? What is the impact of features extracted from the neural network on classification accuracy or recognition results of activities? How to improve the classification accuracy? [17], [18], [19], [20], [21]. 2) In some datasets, the accuracy of classifying activity applied neural networks are not high enough and there is room for improvement, but the accuracy cannot be improved by the conventional methods such as tuning. At the same time, in some datasets, the classification accuracy of different activities is imbalanced[22][23][24]. Moreover, the accuracy of classifying some activities is relatively high, however, the accuracy on other activities is relatively low. These activities of low-accuracy classification becomes a factor that restricts the improvement of accuracy on the whole datasets[25][26]. So how to improve the recognition accuracy of low-accuracy activities?

---

[†] The first two authors contributed equally to this work

To solve the above problems, we should try to understand the internal mechanism of neural network and the impact of sensor data and features on the classification results. Therefore, we visualized the features based on activity data. Currently, most researchers do not clearly know what the data features are extracted from the deep learning model and few publications on data feature visualization exist. In image processing, they give insight into the function of intermediate feature layers and the operation of classifiers [27]. They also perform an ablation study to discover the performance contributions of different model layers. The work in [28] makes it possible to learn about multiple layers of representation and we show models with four layers. According to the activated feature map in the visual convolution neural network, and the signal intensity of RSSI data, Liu, X., Liu, L., analyzed the important characteristics of the extraction phase in the execution stage of the emergency room [29]. Although CNN can extract high-level features of sensors well, the meaning of feature representation as extracted by the convolution kernel remains unclear, so it is necessary to analyze by visualization. Moreover, the past work did not use visual results to affect the classification accuracy. Therefore, this paper will visualize data features extracted from CNN, then we know what data will have an obvious effect on the activity in accordance with the results of visualization; we will extract significant data and send new data into the DNN-based fusion mode. Finally, the experiments show that the classification accuracy rate increased to 96.1%.

In the paper, we present visualizing and understanding CNN-based activity recognition. In summary, this paper makes three contributions. First, we visualized abstract sensor-based activity data features. In-depth analysis of the relationship between sensor data and activities and the impact of features extracted from the neural network on classification results. The extracted features in the deep learning model are no longer a black box and help us to understand the mechanics inside neural networks. It also helps to understand activity and sensor data. Second, we proposed a DNN-based fusion model to improve the accuracy of classification, especially to improve the accuracy of these activities classified in low-accuracy. Third, we presented a promising method to solve the problem that deal with activities classified in low-accuracy. It provides ideas for further study in this field.

## 2  Visualizing and Understanding CNN on dataset HARUSP

The overall flow of experiments is: Train CNN to generate data features; Visualize features to generate feature graph; Analyze and understand features to get relationships between activities and features; Improvement the performance of classification by the fusion Model. As shown in **Figure 1**.

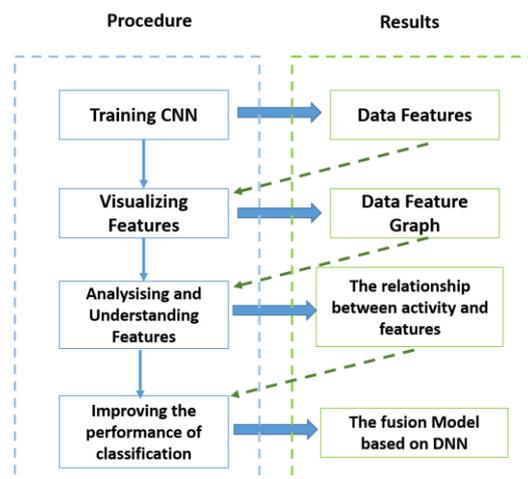

**Figure 1.**Experimental procedure

## 2.1 Dataset

Because this paper is just for providing ideas to readers, we have only carried out experiments on one dataset. We selected a common dataset in the field of HAR: HARUSP; its basic information is shown in **Table 1**.

**Table 1.** The basic information of dataset

| Website | Device | Sensor | Volunteers | Activity |
|---|---|---|---|---|
| http://archive.ics.uci.edu/ml/datasets/Human+Activity+Recognition+Using+Smartphones | smartphone | accelerometer and gyroscope | 30 | 6 types of activities: walking, walking_upstairs, walking_downstairs, sitting, standing, laying |

The following points require explanation:
1) The data matrix of this dataset is composed of nine columns, and they represent the body acceleration *x, y, z* axes, gyroscope *x, y, z* axes, and total acceleration *x, y, z* axes, respectively.
2) The values of the $0^{th}$, $1^{st}$, and $2^{nd}$ columns is mainly affected by the displacement of the human body. The $3^{rd}$, $4^{th}$, and $5^{th}$ columns are mainly influenced by the human body rotation, and the $6^{th}$, $7^{th}$, and $8^{th}$ columns mainly represent the total acceleration. It is assumed that when the smartphone is stationary, it is also under gravitational acceleration of value *1g*, so the acceleration data obtained from the sensor contains the gravity acceleration and the human body's acceleration as the total acceleration.
3) In **Figure 2(a)**, we can know that the smartphone is tied around the waist. **Figure** (b) is the accelerometer's direction in the three axes and **Figure** (c) is the gyroscope's direction in the three axes.

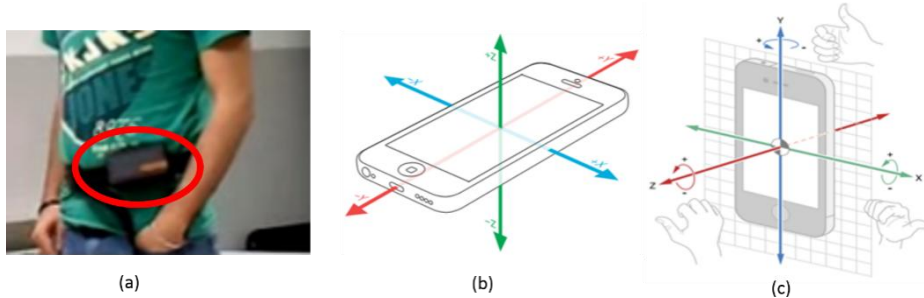

**Figure 2.** Placement map of smartphone and its sensors

## 2.2 CNN-Based feature extraction on dataset HARUSP

### 2.2.1 CNN Model

**Figure** shows the procedure of processing raw sensor data in CNN. First, the raw data is preprocessed and put into the CNN, 2D convolution is carried out, and then the ReLU activation function is used for nonlinear processing. After each convolution layer, we use the max pooling layer to further extract significant features from the sensor data. After high level data feature extraction in the convolution and pooling layer, adding dropout and batch normalization in the network structure can prevent over-fitting. The output of the convolution layer and pooling layer represent advanced features of the input sensor data.

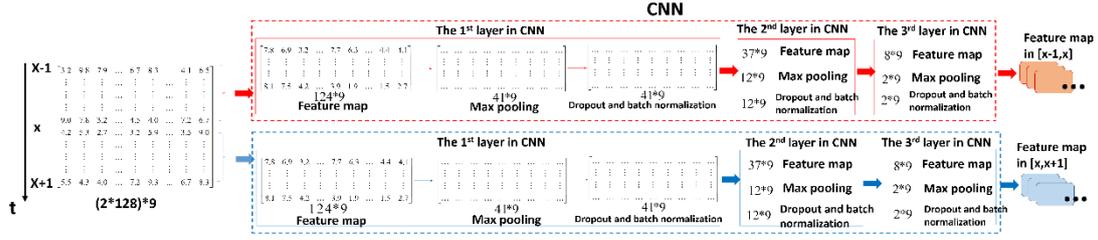

**Figure 3.** The procedure of processing raw sensor data in CNN

**Table 2** shows the basic parameters needed in CNN. It should be noted that these parameters are set based on experiential knowledge.

**Table 2.** The basic parameters needed in CNN

|  | HARUSP |
|---|---|
| The size of input vector | 128*9 |
| The number of kernel | 50 |
| Convolution kernel | 5*1 |
| Pooling size | 3*1 |
| The Probability dropout | 0.5 |
| Learning Rate | 0.01 |
| The number of iterations | 50 |
| The number of samples for each iteration | 32 |

### 2.2.2 Training CNN

The process of training CNN requires the following steps：
1) Initialize all filters and parameters/weights with random numbers;
2) The network takes the sensor data window as input, performs forward propagation (including convolution, ReLU function, max pooling, and forward propagation of the full connection layer), and calculates each class' corresponding output probabilities. Suppose that the output probability at this time is [0.1, 0.5, 0.1, 0.3]; the weights of the first training sample are all random, so this output probability is also similar to the probability of random output.
3) Calculate the total error of the output layer. This paper uses the cross-entropy loss function:

$$C = -\frac{1}{N}\sum_x \left[ y \ln a + (1-y)\ln(1-a) \right] \quad (1)$$

$y$ is the actual output and $a$ is the output value of the neuron. When the real output $a$ is close to the expected output $y$, the cost function is close to *0*.
4) The backpropagation algorithm calculates errors relative to all weights gradient, and uses the gradient descent algorithm to update all weights and parameters in networks, thus minimizing the loss function.

When the sensor data is input again, the probability of the output may be [0.1, 0.1, 0.7, and 0.1], which is closer to the goal of [0, 0, 1, and 0]. Therefore, the process of training CNN includes constantly updating weights and parameters.

### 2.3 Visualizing and Understanding CNN-Based Activity Recognition on HARUSPDataset

The following two points need to be explained:
1) We build two-dimensional feature graphs in which the longitudinal axis represents the time series in the time window; i.e., each row represents data at a time, and the horizontal axis represents the data of nine attributes collected by sensors.
2) We performed the visualization of each type of activity layer-by-layer from a one-layer CNN to a

five-layer CNN. According to this experiment, we can see the influence of the CNN layer on the feature extraction and each activity data feature graph in each layer is composed of one time window selected randomly by the system. Then, we did three-layer CNN extraction feature visualization for each activity type. According to this experiment, we can see which data features can significantly influence the activity, and each activity data feature graph is composed of six time windows that are selected randomly by the system.

### 2.3.1 Visualization of activity data feature graph extracted layer-by-layer

Different features are extracted from different numbers of layers in CNN, in order to better understand the effect of different numbers of layers, we visualize the features extracted from one-layer to five-layer. **Figure** shows six kinds of activity data feature graphs extracted from CNN layer-by-layer. Overall, all six kinds of activity data feature graphs on the third layer are more obvious and more discriminative. Meanwhile, different activities have different main characteristics that facilitate the distinguishing of classifiers. Moreover, we do experiments about the CNN structure whose layers are from 1 to 5 on HARUSP. As you can see from **Figure** , the accuracy and F1-score of the three-layer CNN structure are the best. Therefore, we study the features of the three-layer in depth.

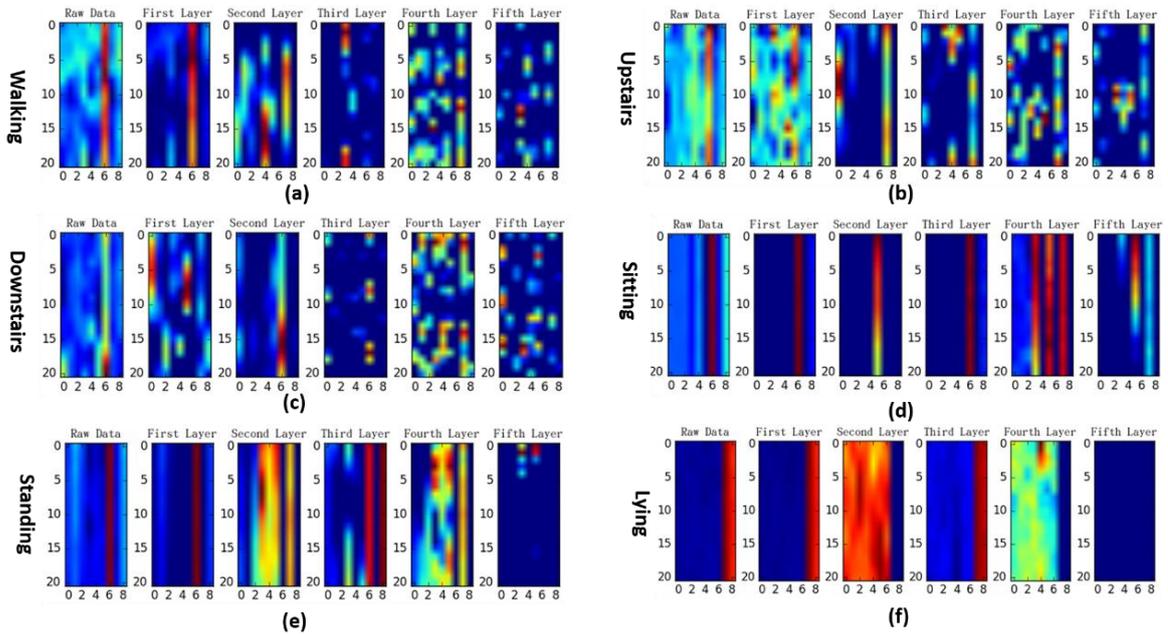

**Figure 4.** Six activity data feature graphs extracted layer-by-layer from CNN

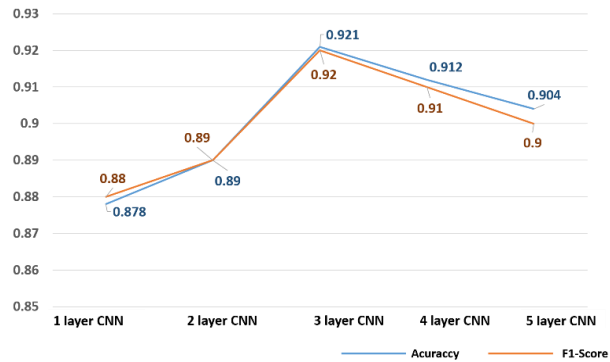

**Figure 5.** The performance of the CNN structure with 1-layer to 5-layer on HARUSP

### 2.3.2 Visualization of activity data feature graph extracted from three-layer CNN

We studied in detail the experiment scenarios and experiment setup for this dataset and analyzed the impact of activity on sensor data according to the experimentally recorded video to help usunderstand the relationship between features and activity. **Figure** shows the feature graphs of the six types of activity after convolution of three-layer CNN. And Table 3 shows the detail description of Figure6.

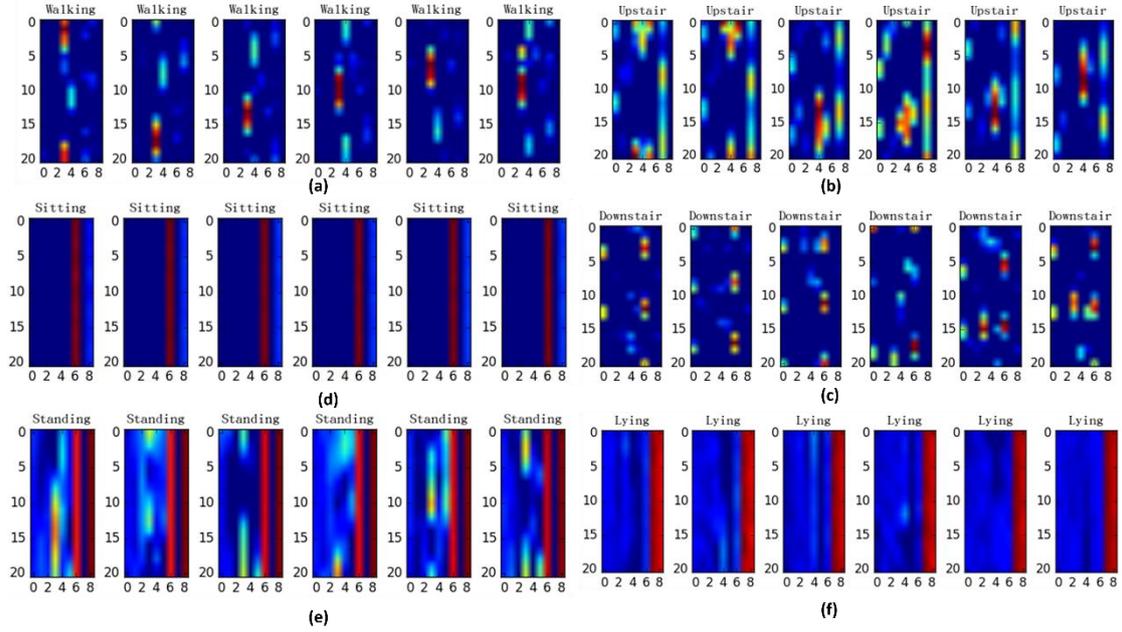

**Figure 6.** Six activity data feature graphs extracted from a three-layer CNN

Table 3. The detail description of Figure6

| activity | Analysis combined with the real life scenarios | The number of columns of the significant features in the feature map | Whether the actual analysis and the visualization result are consistent |
|---|---|---|---|
| walking | People walk no longer in a straight line--->Affect the x-axis of Figure 2 (c); Phone will rotate slightly around y axes--->Affect the y-axis of Figure 2 (c) | the 3th column in Figure 6 (a); the 4th column in Figure 6 (a) | consistent |
| Upstair | The body's displacement will happen in the upper and lower range--->Affect the x-axis of Figure 2 (b); The weight of gravity acceleration when going upstairse--->Affect the y-axis of Figure 2 (b) | the 0th column in Figure 6 (b); the 7th column in Figure 6 (b) | consistent |
| Downstair | The body's displacement will happen in the upper and lower range--->Affect the x-axis of Figure 2 (b); The weight of gravity acceleration when downstairs--->Affect the x-axis of Figure 2 (b) | the 0th column in Figure 6 (c); the 6th column in Figure 6 (c) | consistent |
| Sitting | Affected by the acceleration of gravity--->Affect the x-axis of Figure 2 (b) | the 6th column in Figure 6 (d) | consistent |
| Standing | Slight shaking of the human body when | the 3th column in Figure | consistent |

| | | |
|---|---|---|
| | standing--->Affect the x-axis of Figure 2 (c); | 6 (e); |
| | Affected by the acceleration of gravity--->Affect the x-axis of Figure 2 (b) | the 6th column in Figure 6 (e) |
| Lying | Affected by the acceleration of gravity--->Affect the z-axis of Figure 2 (b) | the 8th column in Figure 6 (f) consistent |

According to the above detailed analysis, each activity has its own significant features columns. we can see these columns have the greatest impact on the recognition of activities. (see**Table** 4).

Table 4.The relationship between column and activity

| | Walking | Upstairs | Downstairs | Sitting | Standing | Lying |
|---|---|---|---|---|---|---|
| Features columns | 3,4 | 0,7 | 0,6 | 6 | 3,6 | 7,8 |

We can further verify the correctness of the above analysis experimentally. The verification method is occlusion[27]; the specific practice is covering a column of data (setting the value as 0), then checking the influence on the effect of classification. Finally, we determine which data has an impact on various types of activity. We used the above method to verify each of the activities on the HARUSP dataset, as shown in **Table 5**.Take Sitting as an example, when covering the 0th columns, 99% of the samples will still be classified into walking, indicating that the 0th column is not the main feature of walking activities. When covering the 6th column, only 9% of samples will be classified as walking, indicating that the sixth columns are the main characteristics of walking activities. The same way as other activities are verified, we mark the column that each activity involves as yellow. As you can see, the results of **Table** 4 are essentially consistent with those of**Table 5**.

Table 5. The experiment result of occlusion method

| | Walking | Upstairs | Downstairs | Sitting | Standing | Lying |
|---|---|---|---|---|---|---|
| **Sample number** | 477 | 435 | 364 | 411 | 475 | 535 |
| **The 0th Column** | 99% | **28%** | **4%** | 99% | 99% | 100% |
| **The 1st Column** | 98% | 98% | 100% | 96% | 100% | 100% |
| **The 2nd Column** | 99% | 96% | 100% | 95% | 100% | 100% |
| **The 3rd Column** | **19%** | 84% | 76% | 100% | **39%** | 99% |
| **The 4th Column** | **54%** | 96% | 89% | 83% | 100% | 100% |
| **The 5th Column** | 60% | 97% | 94% | 97% | 99% | 100% |
| **The 6th Column** | 94% | 97% | 95% | **9%** | **52%** | 100% |
| **The 7th Column** | 96% | **46%** | 100% | 90% | 86% | **89%** |
| **The 8th Column** | 99% | 96% | 96% | 96% | 98% | **92%** |

In this chapter we visualized CNN-based activity recognition on the HARUSP dataset. Through visualization, the sensor data becomes intuitive. And then we get a sufficient understanding of activities, sensor data, and the mechanisms inside neural networks. Because of the different types of activities, the effects of activities on the sensors are different, and the collected sensor data has different characteristics. Neural networks extract different features based on these sensor data. For each activity, some of the features have no effect on the recognition of the activity, but some of the features have a great impact on the recognition of the activity, so different activities have their own significant features. At the same time, the neural networks mainly identify activities based on these significant features.

In the dataset, misclassification was mainly due to standing and sitting. The main reason is that these two types of activities have the same features (the 6th column), representing the effect of the acceleration of gravity. The difference is that when the human body stands there is a slight rotation, the standing activity has the features of the third column, but the features is not obvious because of the small rotation of the human body. As a result, the classification error rate of the two activities is relatively high. After understanding of the visualization features, we proposed a fusion model to improve the accuracy.

## 3 The DNN-based fusion model and Results

Based on the above visualization, we clearly understand the relationship between activity and sensor data, and we find that each activity has corresponding significant features. Activity recognition is mainly based on the significant features, while other features have little effect on the classification. So we extracted the significant features related to the activity and used these features to train the model. Because of the use of the significant features and the exclusion of extraneous features, the model can better recognise difficult-to-distinguish activities and improve the recognition accuracy of all activities. Therefore, in the experiment, we use the main features to train two models, and use all features to train one model, and then use the voting mechanism to fuse the classification results of these three models to improve the recognition accuracy.

According to the above analysis, the two activities that are difficult to distinguish are sitting and standing. The feature columns related to these two activities are the *3rd, 4th, 5th, 6th, 7th, and 8th*, of which the *3rd, 4th, 6th, and 7th* columns are more influential. Therefore, in the experiment, we trained three models:1) used all the data in the dataset; 2) extracted the *3rd, 4th, 5th, 6th, 7th, and 8th* columns of data; 3) extracted the *3rd, 4th, 6th, and 7th* columns of data for training, and then used the voting mechanism to fuse the results of the three models.

**Figure** shows the training procedure of our model. During the training process, the training data were copied into three copies, they are respectively *Train Data1, Train Data2,* and *Train Data3,* respectively. *Train Data1* is sent directly into the *Model-M* for training, in which the *Model-M* structure is three-layer CNN and one-layer LSTM; CNN is responsible for extracting high-level activity data features, LSTM is mainly used to describe the correlation between the activity data time window, the final classification in the Softmax layer. The data of the *3rd, 4th, 5th, 6th, 7th,* and *8th* columns in the *Train Data2* were sent into the *Model-m1* for training, in which the structure of *Model-m1* was the same as that of *Model-M*. The data of *3rd, 4th, 6th,* and *7th* columns in the *Train Data3* are sent into the *Model-m2* for training, in which the structure of *Model-m2* is the same as that of *Model-M*.

**Figure** shows our model's testing procedure. In the process of testing, each piece of data in the test data is copied into *Test Data1, Test Data2,* and *Test Data3*. *Test Data1* is sent into *Model-M* to get the test results of Classification *results-M*, *Test Data2* is sent into *Model-m1* to get the test results of Classification *results-m1*, and *Test Data3* is sent into *Model-m2* to get the test results of Classification *results-m2*. The three classification results then use the voting mechanism to produce the final classification result. If there are two or three consistent results, the final classification results are the consistent results. If the three classification results are not the same, then the final classification results are Classification *results-M*.

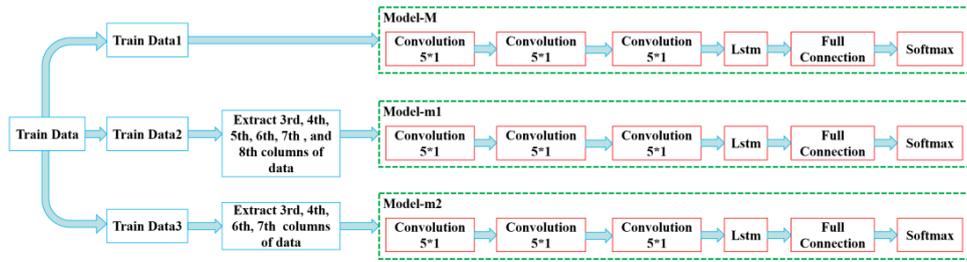

**Figure 7**.Training procedure of the DNN-based fusion model

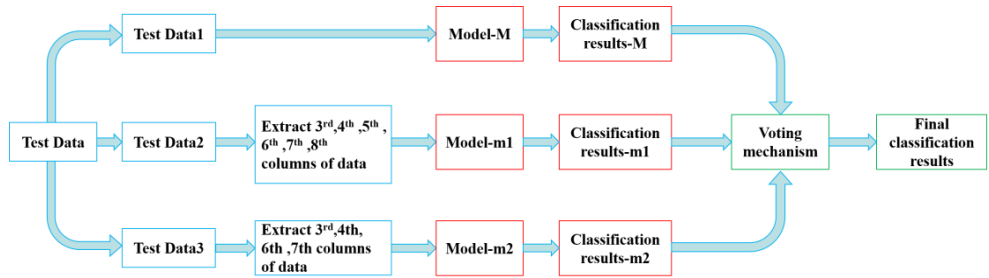

**Figure 8**. Classifying procedure of the DNN-based fusion model

The results are shown in**Table 6**.

Table 6. Comparision of performance of several models

|  | CNN | LSTM | Paper[34] | Paper[35] | Paper[36] | CNN-LSTM | Our method |
|---|---|---|---|---|---|---|---|
| **Accuracy** | 0.919 | 0.892 | 0.89 | 0.893 | 0.962 | 0.951 | 0.961 |
| **F1-Score** | 0.92 | 0.89 | n/a | n/a | n/a | 0.95 | 0.96 |

Next, we compared the confusion matrix of the two methods, because this paper focuses on the deep learning model, so we selected a high recognition rate method (CNN-LSTM) for comparison with our method. The confusion matrixes are listed in **Figure** ; the activities in a row represent predicted activities and the activities in a column are real activities. Upon comparison, it is found that the activities our method improves upon compared to the ordinary CNN-LSTM are mainly the two activities of *Sitting and Standing*; this gave us great inspiration. We found that this method is effective at distinguishing undistinguishable activities. However, since this is just the result of manual extraction, we will propose a method of automatically extracting sensor-based activity data features in future work, so that the new features can improve the accuracy of activity recognition by enhancing the data model.

The confusion matrix of CNN-LSTM

|  | Walking | Upstairs | Downstairs | Sitting | Standing | Lying |
|---|---|---|---|---|---|---|
| Walking | 493 | 3 | 0 | 0 | 0 | 0 |
| Upstairs | 3 | 466 | 2 | 0 | 0 | 0 |
| Downstairs | 3 | 18 | 398 | 0 | 1 | 0 |
| Sitting | 0 | 1 | 0 | 421 | 69 | 0 |
| Standing | 0 | 0 | 0 | 42 | 490 | 0 |
| Lying | 0 | 0 | 0 | 0 | 1 | 536 |

The confusion matrix of our method

|  | Walking | Upstairs | Downstairs | Sitting | Standing | Lying |
|---|---|---|---|---|---|---|
| Walking | 487 | 6 | 3 | 0 | 0 | 0 |
| Upstairs | 1 | 468 | 2 | 0 | 0 | 0 |
| Downstairs | 0 | 13 | 407 | 0 | 0 | 0 |
| Sitting | 0 | 2 | 0 | 431 | 58 | 0 |
| Standing | 0 | 0 | 0 | 29 | 503 | 0 |
| Lying | 0 | 0 | 0 | 0 | 0 | 537 |

**Figure 9**. The two confusion matrixes

## 4  Conclusion and Outlook

This paper presented a novel idea for sensor-based activity recognition. We introduced the application of activity recognition in Deep Neural Network model and the visualization of data features in deep learning. Furthermore, it is most important that we visualize sensor-based activity data features extracted from CNN and the data features are abstract for a large number of researchers. This article is intended to provide a way of thinking for scholars, so we made an in-depth analysis of the visual feature maps on a single dataset HARUSP, which includes the visualization of activity data feature graph extracted from layer-by-layer and three-layer CNN. After the analysis, we manually extracted important data characteristics to our model that have a greater impact on activities. We then verified the correctness of manual feature extraction from two aspects of theory and experiment. The classification accuracy was increased to 96.1%. Experiments revealed that this paper can help solve the problems in Chapter 2.1. Therefore, it is verified by experimentation that our proposed idea is promising, which lays the groundwork for subsequent precision classification.

Through the in-depth analysis of the features, we have identified the relationships among activities, sensor data, and features. In addition, we analyzed how Deep Neural Network model recognize activities based on these features. The significant features mentioned in the experiment can be used not only in the method of the paper but also in other methods and ideas, which has great significance to the improvement of activity recognition. Through visualization, we have a deeper understanding of the mechanism inside the neural network. The ideas, findings and conclusions in the experiment are applicable not only to the field of activity recognition but also to other areas of deep learning.

However, in practice, the manual extraction of features is difficult to verify, particularly in the case of theoretical verification. It is necessary to know the location of sensor when each activity occurs. In future work, we will first present a method for extracting main features automatically. In addition, we ensure the accuracy of the enhancement by considering that we present a fusion model that combines the main features with the original features to enhance the data.